\documentclass[letterpaper]{article} 
\usepackage{aaai2027}  
\usepackage[hyphens]{url}  
\usepackage{graphicx} 
\usepackage{natbib}  
\usepackage{caption} 
\usepackage{algorithm}
\usepackage{algorithmic}
\usepackage{multirow}
\usepackage{amsmath}
\usepackage{amssymb}
\usepackage{newfloat}
\usepackage{listings}
\DeclareCaptionStyle{ruled}{labelfont=normalfont,labelsep=colon,strut=off} 
\floatstyle{ruled}
\newfloat{listing}{tb}{lst}{}
\floatname{listing}{Listing}

\usepackage{booktabs}

\usepackage{hyperref}
\usepackage{navigator}
\nocopyright 

\title{FUSED: Forensic–Semantic Mixture-of-Experts for AI Inpainting \\Detection and Localization}
\author{
Anton Nuzhdin,
Marcel Worring,
Ivona Najdenkoska
}

\affiliations{
Informatics Institute, University of Amsterdam\\
Amsterdam, The Netherlands \\
\texttt{anton.nuzhdin@uva.student.nl, }
\texttt{m.worring@uva.nl, i.najdenkoska@uva.nl} 
}

\begin{document}

\maketitle

\begin{abstract}
Diffusion-based inpainting models modify only a localized part of an image, while many AI-image detectors rely on global artifacts and do not localize. These artifacts vary across generators, limiting detector transfer under distribution shifts. Recent work shows that restoring the authentic pixels outside the inpainted region removes these cues and can degrade pretrained detectors. To address this, we present FUSED, a unified framework for the joint detection and localization of AI-generated inpainting. FUSED combines low-level forensic cues with high-level semantic features using a sparsely-gated Mixture-of-Experts architecture, enabling the model to adaptively prioritize the most relevant signal for each token. For each input, FUSED predicts both an image-level manipulation score and a pixel-level mask of the inpainted area. On the OpenSDID cross-generator benchmark, FUSED achieves the best average detection and localization, with the largest gains on unseen generators. The same model transfers directly to the held-out AutoSplice and CocoGlide benchmarks, more than doubling localization performance. Evaluating each held-out benchmark with and without the global generator artifact further shows that all evaluated methods, ours included, partly read the artifact as evidence of manipulation, and FUSED remains the strongest under both conditions. Code and pretrained models are available at \url{https://github.com/AntonNuzhdin/FUSED}.
\end{abstract}

\section{Introduction}
\label{sec:intro}

Detecting and localizing AI-generated content has become a central
problem in image forensics. Diffusion-based image inpainting has made AI-generated content rapid, accessible, and visually
convincing~\citep{ldm, sd3, sdxl, flux, qwen-image}. In contrast to fully
synthetic generation, AI inpainting modifies only a localized region of an image,
removing, inserting, or replacing semantically important content while maintaining
the realism of the surrounding areas~\citep{verdolina}. Consequently, a forensic
system designed to detect such edits must address two interrelated questions:
\emph{has} the image been manipulated, and \emph{where} is the manipulation located.
\begin{figure}[!tb]
  \centering
  \includegraphics[width=0.9\columnwidth]{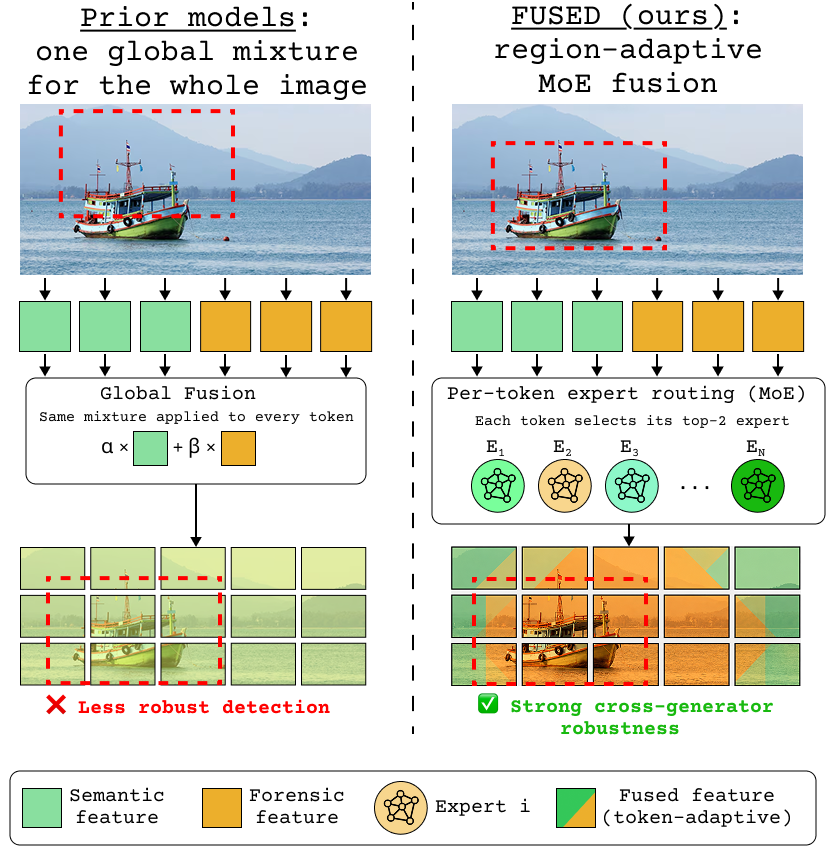}
  \caption{Prior forensic-semantic detectors combine forensic and semantic features using a global image-level fusion. FUSED instead routes each token to a sparse set of experts, enabling token-level fusion and improving robustness to unseen generators.}
  \label{fig:teaser}
\end{figure}
Most detectors take a shortcut around the detection vs localization problem. Specifically, detectors tend to rely on the global cue that distinguishes real from synthetic images in their training data, such as JPEG compression and resolution~\citep{Grommelt}, content and format biases~\citep{bfree}, or frequency misalignment~\citep{dual-data-alignment}. In inpainting, the VAE trace is exactly such a cue: latent diffusion models process the entire image through a variational autoencoder (VAE)~\citep{vae}, leaving a spectral trace across the whole frame, including the background~\citep{inpx}.
Recent work, such as the Inpainting-Exchange intervention (INP-X)~\citep{inpx}, investigates this by showing that restoring the authentic pixels outside the edited region reduces strong detectors to chance. Detection must, therefore, rest on the synthesized region itself.

Reliance on shortcuts becomes unsustainable under transfer, where new
generators arrive without retraining, and there the two kinds of evidence
fail in complementary ways. Low-level forensic cues localize accurately but
generalize poorly across generators, while high-level semantic features
survive the generator change yet cannot point to the synthesized pixels, so
localization suffers even when detection holds. Combining the two is by now
standard~\citep{trufor, aide, cospy}, from feature
concatenation~\citep{aide} to adaptive per-image weighting~\citep{cospy}.
In all of them, a single blend serves the whole image, while the inpainting is
local, so the mixture must instead be resolved token by token
(Figure~\ref{fig:teaser}).

\begin{figure*}[!t]
    \centering
    \includegraphics[width=1.0\textwidth]{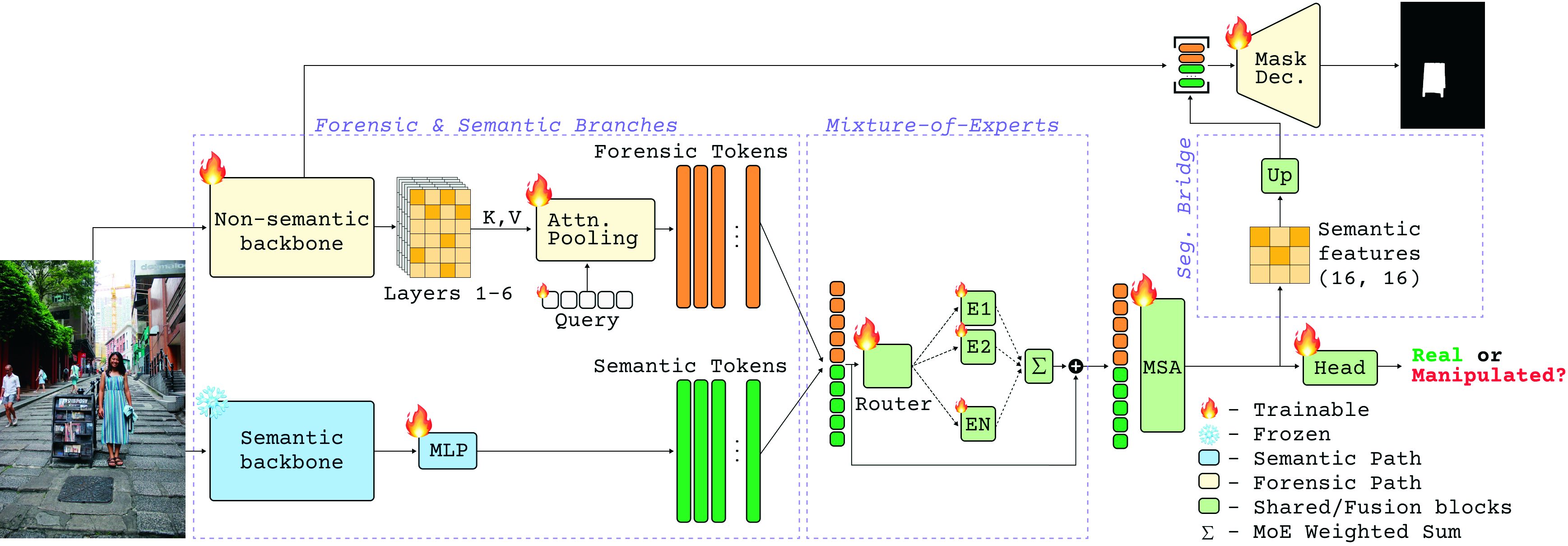}
    \caption{The FUSED architecture. In contrast to the single image-level fusion illustrated in Figure~\ref{fig:teaser}, FUSED represents the forensic and semantic streams as token sets, concatenates them into a joint sequence, and routes each token through a sparse Mixture-of-Experts before cross-stream self-attention. A classification head predicts the image-level manipulation score, while a segmentation bridge combines the forensic feature map with the post-fusion semantic tokens to decode the tampering mask.}
    \label{fig:fused}
\end{figure*}

In this paper, we introduce \textbf{FUSED}, a \textbf{F}orensic-semantic
\textbf{U}nified \textbf{S}parse-\textbf{E}xpert \textbf{D}etector for the joint
detection and localization of AI inpainting (Figure~\ref{fig:fused}). A trainable forensic branch suppresses semantic
content and extracts low-level manipulation traces, while a frozen
semantic branch provides object-level context that remains
stable across generators. A sparsely-gated Mixture-of-Experts (MoE)~\citep{moe} fuses the streams by routing each
token to selected experts. Finally, a segmentation bridge decodes the tampering mask jointly from
the forensic map and the post-fusion semantic tokens, extending localization supervision to the fusion and routing modules.

To evaluate our method, we conduct a series of transfer experiments under three increasingly severe distribution shifts from the training source. Following the OpenSDID protocol~\citep{opensdi}, all methods are trained on images by a single image generator and tested on five generators of the same latent-diffusion family, four of which are unseen. The same checkpoints are then evaluated, without any target-domain training, on AutoSplice~\citep{autosplice} and CocoGlide~\citep{trufor}, two independently constructed benchmarks whose pixel-space editors never pass the image through a latent VAE. Consequently, no global latent-VAE artifact is present, and successful transfer requires identifying evidence within the manipulated region itself. The INP-X benchmark~\citep{inpx} provides the same manipulations with the artifact present and removed, allowing us to isolate how much each transferred decision depends on this cue. Together, these evaluations test cross-generator generalization, transfer across editing pipelines, and robustness to the removal of global artifacts. All three, however, start from a single training generator, leaving open both whether the gains of FUSED persist under heterogeneous training and whether the model remains effective against more recent high-fidelity generators. We therefore also retrain FUSED and MaskCLIP on the large-scale multi-generator So-Fake-Set~\citep{huang2026sofake} and evaluate them on So-Fake-OOD, which contains forgeries from held-out commercial generators.

In summary, our contributions are: (1)~We propose FUSED, which couples a forensic branch with a semantic one through a sparsely-gated
Mixture-of-Experts acting as a cross-stream, per-token fusion operator, and a joint
segmentation bridge that extends localization supervision to the fusion itself.
(2)~We demonstrate state-of-the-art performance in detection and localization from a single training source across generator families and under both artifact conditions. Furthermore, we show that the advantage persists under large-scale multi-generator training and on a contemporary OOD benchmark containing held-out commercial generators.

\section{Related Work}
\label{sec:related}

\paragraph{Detecting AI-generated images.}
\label{subsec:detection}
Most detectors classify an image as real or generated based on the statistical
fingerprints of the generator. Early work exploits the spatial and spectral
artifacts of CNN generators~\citep{cnndetection}, later refined through
up-sampling and neighboring-pixel cues~\citep{npr}, frequency-domain
analysis~\citep{freqnet,gramnet}, or any-resolution spectral
learning~\citep{spai}. Reconstruction-based detectors read diffusion images
off the generative pipeline itself~\citep{dire, aeroblade, latte}, while a
parallel line repurposes CLIP features as a generic real/fake
space~\citep{unifd,rine}. Because either cue alone has blind spots, recent
detectors fuse them: AIDE~\citep{aide} concatenates patch statistics with a
semantic representation, and CO-SPY~\citep{cospy} pairs a semantic stream
with VAE reconstruction residuals. All decide at the image level with a
single blend of evidence, and none localizes, whereas FUSED resolves the
mixture per token and grounds its decision in a predicted mask.

\paragraph{Image manipulation detection and localization.}
\label{subsec:imdl}
The setting of our work is image manipulation detection and localization, in which a method produces both an image-level forgery decision and a per-pixel tampering map. Most approaches analyze low-level forensic traces, revealing tampering through anomaly modeling~\citep{mantranet}, noise and edge streams~\citep{mvssnet}, compression artifacts~\citep{catnet}, spatio-channel correlation~\citep{psccnet}, object-level transformers~\citep{objectformer}, or plain vision transformers~\citep{imlvit}, while TruFor~\citep{trufor} fuses an RGB view with a learned noise fingerprint. Most of these methods obtain their low-level view from handcrafted
extractors, whereas
SparseViT~\citep{sparsevit} shows that a sparse attention design learns
comparable non-semantic features without them. A growing subset targets diffusion inpainting directly, emphasizing the localization map through uncertainty-guided learning~\citep{tanet}, multi-feature fusion~\citep{mfinet}, contrastive objectives~\citep{cflnet}, edge guidance~\citep{ecnet}, or conditional diffusion used as a localizer~\citep{inpdiffusion}. Seeking more robust representations, DeCLIP~\citep{declip} and MaskCLIP~\citep{opensdi} instead read edits from CLIP~\citep{clip} features, which transfer across generators more effectively; yet their localization declines even where detection remains effective, as our experiments show in Section~\ref{subsec:transfer_results}.

A separate line introduces mixture-of-experts routing into forensics.
MoE-FFD~\citep{moeffd} and MoNFAP~\citep{monfap} route among adapter
or noise experts for face manipulation, Forensic-MoE~\citep{forensicmoe}
pre-trains adapter experts on different synthetic sources, and
OmniAID~\citep{omniaid} routes among content-domain semantic experts
alongside a fixed artifact expert. These methods target fully synthetic
or face-manipulated imagery, their experts operate within a single
representation, and specialization is largely imposed by pre-training
each expert on a predefined data partition. FUSED instead trains
end-to-end on a single undivided set, letting the router organize the
experts as a cross-stream, per-token fusion operator between two
heterogeneous encoders, so we treat these methods as design references
rather than baselines.

\paragraph{Shortcut learning in AI-image detection.}
\label{subsec:robustness}
A recurring finding is that AI-image detectors succeed for the wrong reasons. A line of work documents dataset biases that detectors latch onto, such as JPEG compression~\citep{Grommelt}, and proposes bias-free training~\citep{bfree} or cross-source data alignment~\citep{dual-data-alignment} to counter them. For inpainting specifically, the INP-X dataset~\citep{inpx} isolates the effect by removing the global VAE artifact while preserving the synthesized content, after which the accuracy of pretrained detectors collapses, suggesting that robust inpainting forensics should be content-aware rather than relying on a global real/fake decision. How to stay robust to the exchange while keeping detection and localization strong together remains open. We take this question to the transfer setting and measure it directly.

\section{Methodology}
\label{sec:method}


\paragraph{Overview.}
We address image manipulation analysis as a joint detection-and-localization task. Localization relies on non-semantic, manipulation-sensitive evidence, whereas robust authenticity detection benefits from combining this evidence with high-level semantic context. Our model, FUSED (Figure~\ref{fig:fused}), follows this division of labor: a trainable non-semantic encoder learns manipulation-sensitive representations, while a frozen semantic backbone provides object-level context. To enable token-wise, cross-stream adaptation, their token representations are fused through a sparsely gated MoE. Given an input $x\in\mathbb{R}^{3\times512\times512}$, FUSED produces an authenticity logit $\hat{y}$ and a full-resolution tampering mask $\hat{S}$.

\paragraph{Forensic branch.}
The forensic branch must expose low-level manipulation traces while suppressing image content. We instantiate it with SparseViT~\citep{sparsevit}, an image manipulation encoder that learns non-semantic features. It is a four-stage hierarchical transformer that replaces the dense global self-attention of a standard ViT~\citep{vit} with
sparse attention. At each layer, the spatial feature tokens are partitioned into spatially strided, non-overlapping groups, and attention is computed only among the non-adjacent tokens assigned to the same group.
Because image semantics require dense, continuous interactions whereas
non-semantic evidence is content-independent and breaks only where a
manipulation occurs, this sparse design suppresses
semantic content and surfaces manipulation-sensitive, content-independent
features. We call this stream the \textit{forensic branch}, as it supplies the
low-level content-independent evidence. Following the original SparseViT approach, we aggregate its
multi-scale features $\{C_j\}$ with learnable channel-wise weights
$\gamma_j$ and layer normalization into a single fused forensic map $F$:
\begin{equation}
F = \mathrm{LN}\Big(\sum_{j} \gamma_j\, C_j\Big).
\label{eq:eqone}
\end{equation}

We then flatten $F$ into a token sequence and summarize it with an
attention-pooling module that cross-attends a set of $m{=}8$ learnable query latents, yielding a compact forensic token set $T_{svit}\in\mathbb{R}^{m\times d}$. The latents are orthogonally
initialized and kept complementary during training by a diversity penalty $\mathcal{L}_{div}=\frac{1}{m(m-1)}\sum_{i\neq j}\big(\hat{t}_i^{\top}\hat{t}_j\big)^2$ on the off-diagonal cosine similarities of the $\ell_2$-normalized pooled tokens $\hat{t}_i=T_{svit,i}/\lVert T_{svit,i}\rVert$.

\paragraph{Semantic branch.}
The semantic branch must provide an object-level context. We instantiate it with a frozen CLIP-ConvNeXt-XXL~\citep{convnext}, following AIDE~\citep{aide}. We use it as a fixed semantic feature extractor: its spatial output of the final-stage (a $16\times16$ grid of feature vectors) is projected by a $1{\times}1$ convolution to a sequence of semantic tokens $T_{cnx}\in\mathbb{R}^{n_c\times d}$ of width $d$. Freezing this branch keeps the trainable parameter count modest and acts as a regularizer against overfitting to dataset-specific artifacts.

\paragraph{Mixture-of-experts fusion.}
The two branches speak different ``languages,'' and the manipulations they must explain are heterogeneous. Therefore, we route tokens through a sparse MoE layer,
which allows the model to dedicate distinct expert sub-networks to distinct
manipulation regimes. Concatenating the branch tokens into
$Z=[\,T_{svit};T_{cnx}\,]$, we fuse them by applying the MoE layer before
self-attention.
\begin{equation}
\begin{aligned}
Z' &=
\mathrm{LN}\!\left(
Z + \mathrm{MoE}(\mathrm{LN}(Z))
\right), \\
Z'' &=
\mathrm{LN}\!\left(
Z' + \mathrm{MSA}(Z')
\right).
\end{aligned}
\label{eq:moe_fusion}
\end{equation}
Thus, per-token expert specialization is resolved first, and the subsequent self-attention models the cross-branch interactions. For each token, a linear router selects the $k{=}2$ most relevant of $E{=}8$ SwiGLU~\citep{swiglu} experts and mixes their outputs by the renormalized routing weights,
\begin{equation}
\begin{aligned}
\mathrm{MoE}(z)
&=
\sum_{i\in\mathcal{K}}\tilde{r}_i e_i(z),
\qquad
\tilde{r}_i
=
\frac{r_i}{\sum_{j\in\mathcal{K}}r_j},
\\
r
&=
\mathrm{softmax}(W_r z).
\end{aligned}
\end{equation}
where $\mathcal{K}$ indexes the top-$k$ experts. To prevent routing collapse, we add the standard load-balancing term
$\mathcal{L}_{MoE}=E\sum_i f_i P_i$, with $f_i$ and $P_i$ respectively being the
dispatch fraction and mean gate probability of expert $i$.

\paragraph{Prediction heads and segmentation bridge.}
The authenticity logit is obtained by pooling the fused tokens $Z''$
into a classification head. For localization, decoding a mask from the forensic map $F$ captures
local forensic traces but ignores the semantic context of the frozen backbone,
and leaves the MoE fusion supervised only by the image-level label. The
segmentation bridge addresses both issues by decoding the final mask $\hat{S}$
jointly from the two streams: the ConvNeXt slice $Z''_{cnx}$ is reshaped to its
$16\times16$ grid, upsampled to the resolution of $F$, and
concatenated with it along the channel dimension. The concatenated features are passed to a convolutional decoder $\mathcal{D}$, consisting of two $3{\times}3$ Conv--GroupNorm--GELU blocks followed by a $1{\times}1$ projection:
\begin{equation}
\hat{S}=\mathrm{Up}\Big(\mathcal{D}\big([\,F \,;\, \mathrm{Up}(Z''_{cnx})\,]\big)\Big).
\end{equation}
The bridge fuses forensic evidence and semantic context at the feature level. Because the mask is decoded from the post-fusion representation, the localization loss also supervises the fusion, MoE, and semantic-projection modules.

\paragraph{Training details.}
The model is trained end-to-end with localization, boundary, classification, load-balancing, and diversity terms:
\begin{equation}
\begin{aligned}
\mathcal{L}
={}&
\lambda_{\mathrm{seg}}\mathcal{L}_{\mathrm{BCE}}
+
\lambda_{\mathrm{edge}}\mathcal{L}_{\mathrm{EDG}}
+
\lambda_{\mathrm{cls}}\mathcal{L}_{\mathrm{CE}}
\\
&+
\lambda_{\mathrm{MoE}}\mathcal{L}_{\mathrm{MoE}}
+
\lambda_{\mathrm{div}}\mathcal{L}_{\mathrm{div}}.
\end{aligned}
\label{eq:loss}
\end{equation}
$\mathcal{L}_{EDG}$ is the edge-weighted
loss, widely used by previous works~\citep{trufor, opensdi, imlvit}. 
We apply the localization terms only to manipulated images, while authentic images are supervised by the classification term alone.

\section{Experiments and Results}
\label{sec:exps}

\begin{table*}[!t]
\centering
{%
\small
\setlength{\tabcolsep}{4.0pt}
\begin{tabular}{@{}l*{6}{cc}@{}}
\toprule
\multirow{2}{*}{Method}
& \multicolumn{2}{c}{SD1.5}
& \multicolumn{2}{c}{SD2.1}
& \multicolumn{2}{c}{SDXL}
& \multicolumn{2}{c}{SD3}
& \multicolumn{2}{c}{Flux.1}
& \multicolumn{2}{c}{Average}
\\
& IoU & F1 & IoU & F1 & IoU & F1 & IoU & F1 & IoU & F1 & IoU & F1
\\
\midrule
MVSS-Net~\citep{mvssnet}
& 57.9 & 65.3 & 44.9 & 51.8 & 14.7 & 18.5 & 26.9 & 32.7 & 4.8 & 6.4 & 29.8 & 34.9 \\
CAT-Net~\citep{catnet}
& 66.4 & \underline{74.8} & 54.6 & 62.3 & 25.5 & 30.7 & 35.6 & 42.1 & 5.0 & 6.6 & 37.4 & 43.3 \\
PSCC-Net~\citep{psccnet}
& 54.7 & 64.2 & 36.7 & 44.8 & 19.7 & 26.1 & 29.3 & 37.3 & 8.2 & 11.6 & 29.7 & 36.8 \\
ObjectFormer~\citep{objectformer}
& 51.2 & 65.7 & 47.4 & 41.4 & 7.4 & 9.8 & 9.4 & 12.6 & 5.3 & 7.3 & 24.1 & 27.4 \\
TruFor~\citep{trufor}
& 63.4 & 71.0 & \underline{54.7} & 61.9 & 26.6 & 31.9 & 32.3 & 38.5 & 7.6 & 9.7 & 36.9 & 42.6 \\
DeCLIP~\citep{declip}
& 37.2 & 43.4 & 35.7 & 41.9 & 14.6 & 18.2 & 27.3 & 33.4 & 11.2 & 14.3 & 25.2 & 30.3 \\
IML-ViT~\citep{imlvit}
& \underline{66.5} & 73.6 & 44.8 & 50.6 & 21.5 & 26.0 & 23.6 & 28.3 & 6.1 & 7.9 & 32.5 & 37.3 \\
MaskCLIP~\citep{opensdi}
& \textbf{67.1} & \textbf{75.6} & \textbf{55.5} & \underline{62.9} & \underline{31.0} & \underline{37.0} & \underline{43.8} & \underline{51.2} & \underline{16.2} & \underline{20.3} & \underline{42.7} & \underline{49.4} \\
\midrule
\textbf{FUSED (Ours)}
& 59.6 & 70.5 & 52.4 & \textbf{63.9} & \textbf{37.5} & \textbf{48.3} & \textbf{50.0} & \textbf{62.1} & \textbf{22.9} & \textbf{31.8} & \textbf{44.5} & \textbf{55.3} \\
\bottomrule
\end{tabular}
}
\caption{Pixel-level localization on the OpenSDID test generators (per-image pixel F1 and IoU). Bold is best, underline second-best. FUSED achieves the best average localization, with its largest margins on the three most distant generators.}
\label{tab:opensdi_localization}
\end{table*}

\begin{table*}[!t]
\centering
{%
\small
\setlength{\tabcolsep}{4.0pt}
\begin{tabular}{@{}l*{6}{cc}@{}}
\toprule
\multirow{2}{*}{Method}
& \multicolumn{2}{c}{SD1.5}
& \multicolumn{2}{c}{SD2.1}
& \multicolumn{2}{c}{SDXL}
& \multicolumn{2}{c}{SD3}
& \multicolumn{2}{c}{Flux.1}
& \multicolumn{2}{c}{Average}
\\
& F1 & Acc. & F1 & Acc. & F1 & Acc. & F1 & Acc. & F1 & Acc. & F1 & Acc.
\\
\midrule
CNNDetection~\citep{cnndetection}
& 84.6 & 85.0 & 71.6 & 75.9 & 59.7 & 68.7 & 56.3 & 67.1 & 35.7 & 57.6 & 61.6 & 70.9 \\
GramNet~\citep{gramnet}
& 80.5 & 80.4 & 74.0 & 76.7 & 65.3 & 70.8 & 64.4 & 70.3 & 52.0 & 63.4 & 67.2 & 72.3 \\
FreqNet~\citep{freqnet}
& 75.9 & 77.7 & 61.0 & 68.4 & 53.2 & 64.0 & 53.5 & 64.4 & 38.5 & 57.1 & 56.4 & 66.3 \\
NPR~\citep{npr}
& 79.4 & 79.3 & 81.7 & 81.8 & 72.1 & 74.3 & \underline{73.4} & 75.5 & \underline{67.6} & \underline{71.4} & 74.9 & 76.4 \\
UniFD~\citep{unifd}
& 77.5 & 77.6 & 80.6 & 81.9 & 70.7 & 74.8 & 71.1 & 75.2 & 61.1 & 69.1 & 72.2 & 75.7 \\
RINE~\citep{rine}
& 91.1 & 91.0 & 87.5 & 88.1 & 73.4 & 78.8 & 72.1 & 76.8 & 55.9 & 67.0 & 76.0 & 80.3 \\
\midrule
MVSS-Net~\citep{mvssnet}
& 93.5 & 93.7 & 79.3 & 82.3 & 59.9 & 70.4 & 62.8 & 72.1 & 27.6 & 56.8 & 64.6 & 75.1 \\
CAT-Net~\citep{catnet}
& \textbf{96.2} & \underline{96.2} & 79.3 & 82.5 & 64.8 & 73.3 & 65.3 & 73.6 & 22.7 & 55.3 & 65.6 & 76.2 \\
PSCC-Net~\citep{psccnet}
& \underline{96.1} & 96.1 & 76.9 & 80.9 & 55.7 & 68.8 & 59.8 & 70.9 & 51.8 & 67.0 & 68.0 & 76.8 \\
ObjectFormer~\citep{objectformer}
& 71.7 & 75.2 & 66.8 & 72.6 & 49.2 & 62.9 & 48.3 & 62.5 & 37.9 & 58.1 & 54.8 & 66.3 \\
TruFor~\citep{trufor}
& 90.1 & \textbf{97.7} & 35.9 & 55.6 & 58.0 & 66.4 & 59.7 & 67.5 & 49.1 & 61.6 & 58.6 & 69.8 \\
DeCLIP~\citep{declip}
& 80.7 & 78.3 & 84.0 & 82.8 & 70.7 & 70.6 & 69.9 & 68.4 & 51.8 & 65.6 & 71.4 & 73.1 \\
IML-ViT~\citep{imlvit}
& 94.5 & 75.7 & 69.7 & 61.2 & 41.0 & 50.0 & 44.7 & 51.2 & 18.2 & 43.6 & 53.6 & 56.4 \\
MaskCLIP~\citep{opensdi}
& 92.6 & 92.7 & \underline{88.7} & \underline{89.5} & \underline{78.0} & \underline{81.2} & 73.1 & \underline{78.0} & 56.5 & 68.5 & \underline{77.8} & \underline{82.0} \\
\midrule
\textbf{FUSED (Ours)}
& 93.9 & 94.0 & \textbf{95.4} & \textbf{95.5} & \textbf{90.2} & \textbf{90.8} & \textbf{89.2} & \textbf{90.1} & \textbf{71.2} & \textbf{77.2} & \textbf{88.0} & \textbf{89.5} \\
\bottomrule
\end{tabular}
}
\caption{Image-level detection on the OpenSDID test generators (F1 and
accuracy at a fixed threshold of $0.5$). Bold indicates
the best result, and underlining indicates the second-best. FUSED achieves the best average detection.}
\label{tab:opensdi_classification}
\end{table*}

\subsection{Experimental Setup}
\label{subsec:setup}

\paragraph{Datasets, protocols, and metrics.}
For the OpenSDID-centered evaluation, all methods, including FUSED and
every baseline, are trained only on the OpenSDID SD1.5 split and tested
under three increasingly severe distribution shifts. First, following the OpenSDI
protocol, all methods are evaluated on the five
OpenSDID test generators: SD1.5, which is in-domain, and SD2.1, SDXL,
SD3, and Flux.1, which are unseen
(Tables~\ref{tab:opensdi_localization}
and~\ref{tab:opensdi_classification}). Second, all methods are
evaluated on AutoSplice, which contains local edits
produced by DALL-E~2, and CocoGlide, which contains
GLIDE inpaintings of COCO crops (Table~\ref{tab:benchmarks_orig}).
Both editors operate in pixel space, so neither benchmark contains the
global latent-VAE artifacts. Third, the
INP-X benchmark~\citep{inpx}, generated using Kandinsky~2.2,
OpenJourney, and Stable Diffusion~v1.4, holds the manipulation fixed and varies
only the global artifact, providing each inpainted image with the
artifact present, INP, and removed, INP-X
(Table~\ref{tab:inpxtest}). The generators, source data, and exchange
operation are all unseen during training. Beyond these three shifts, we test whether the advantage of FUSED
depends on the narrowness of the training source by retraining FUSED
and MaskCLIP on the multi-generator So-Fake-Set~\citep{huang2026sofake}
and evaluating on the So-Fake-OOD benchmark.

Localization is evaluated using per-image pixel F1 and IoU, and
detection on OpenSDID using F1 and accuracy at a fixed threshold of
$0.5$. OpenSDID averages are unweighted
means over the five generators. On the held-out benchmarks and the
INP-X pair, we additionally report ROC-AUC, which, unlike accuracy at
a fixed threshold, is unaffected by score-distribution shift under
transfer.

\paragraph{Baselines.}
The evaluation includes detection-and-localization methods from both
evidence families. The forensic-trace methods are
MVSS-Net~\citep{mvssnet}, CAT-Net~\citep{catnet},
PSCC-Net~\citep{psccnet}, ObjectFormer~\citep{objectformer},
IML-ViT~\citep{imlvit}, and TruFor~\citep{trufor}. The
foundation-model methods are DeCLIP~\citep{declip} and
MaskCLIP~\citep{opensdi}. Following the OpenSDI
protocol~\citep{opensdi}, we also report detection-only
baselines on OpenSDID~\citep{cnndetection,gramnet,freqnet,npr,unifd,rine}. On So-Fake we compare against MaskCLIP only, as it is the strongest
prior method on OpenSDID and the held-out benchmarks.

\begin{figure*}[!t]
    \centering
    \includegraphics[width=0.9\textwidth]{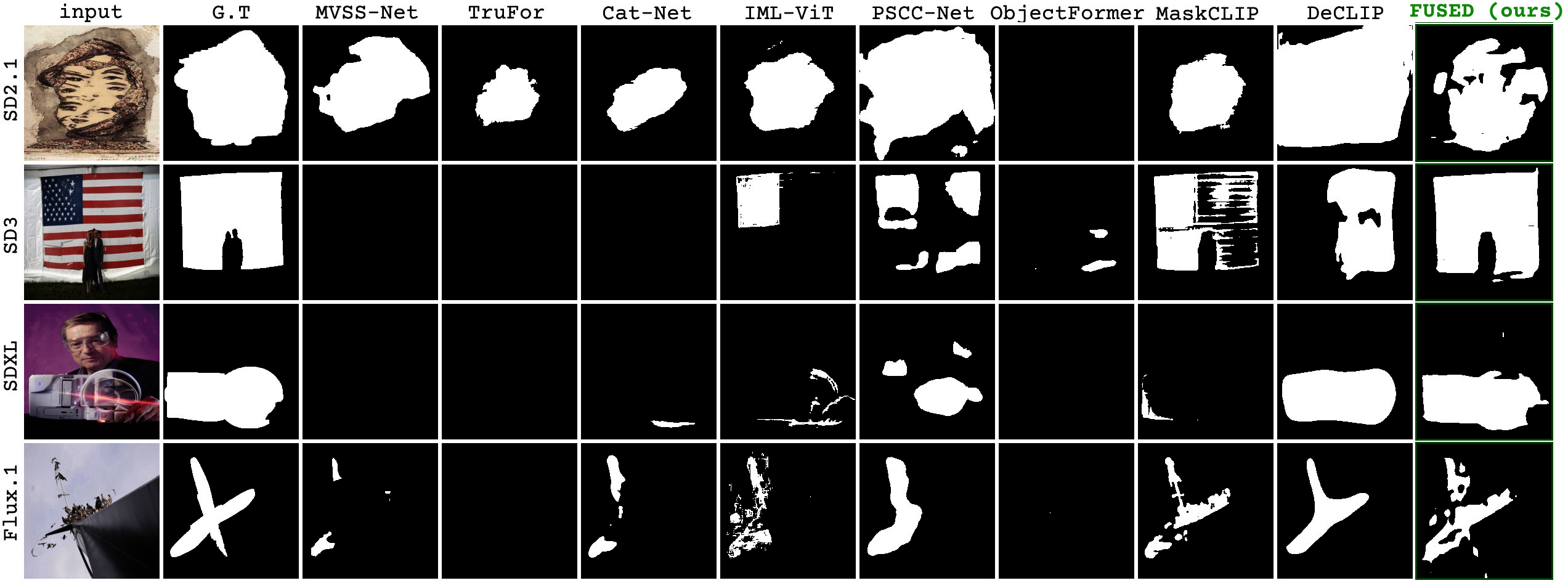}
    \caption{Qualitative localization results on the four unseen OpenSDID generators, one per row. On unseen generators, several baselines produce empty or fragmented masks, while FUSED more consistently localizes the manipulated region.}
    \label{fig:qualitative_opensdi}
\end{figure*}

\paragraph{Implementation details.}
FUSED uses the architecture described in Section~\ref{sec:method}: a
trainable SparseViT forensic branch, a frozen CLIP-ConvNeXt-XXL
semantic branch, $512\times512$ inputs, sparse MoE fusion with
$E{=}8$ experts and top-$k{=}2$ routing, and a joint
forensic-semantic decoder. The loss weights in Eq.~\eqref{eq:loss} are
$\lambda_{\mathrm{seg}}=\lambda_{\mathrm{cls}}=1$, $\lambda_{\mathrm{edge}}=0.5$, and
$\lambda_{\mathrm{MoE}}=\lambda_{\mathrm{div}}=0.01$.

Following the original OpenSDI setting~\citep{opensdi}, all methods are
trained for 20 epochs. FUSED is trained under one
fixed random seed using AdamW with a learning
rate of $3\times10^{-4}$ and an effective batch size of $64$ on two
NVIDIA H100 GPUs (94\,GB each). We use the final training epoch as the evaluation checkpoint. For the So-Fake experiment, the number of training samples varies
substantially across generator sources, so we use square-root sampling,
with the sampling probability of generator $g$ proportional to
$\sqrt{N_g}$, where $N_g$ is its number of training examples. Both
models are trained for two epochs with validation every $10\%$ of the
total training steps, and we select the checkpoint with the highest
validation pixel F1. All other model and optimization settings follow
the OpenSDID experiments.

\begin{table*}[!t]
\centering
{%
\small
\setlength{\tabcolsep}{4.0pt}
\begin{tabular}{@{}lcccccccccc@{}}
\toprule
\multirow{2}{*}{Model}
& \multicolumn{5}{c}{AutoSplice}
& \multicolumn{5}{c}{CocoGlide}
\\
\cmidrule(lr){2-6} \cmidrule(lr){7-11}
& Pix.\ F1 & IoU & F1 & AUC & Acc.
& Pix.\ F1 & IoU & F1 & AUC & Acc.
\\
\midrule
MVSS-Net~\citep{mvssnet}
& 7.7 & 5.4 & 42.6 & \underline{75.2} & \underline{53.7}
& 3.7 & 2.9 & 14.7 & 65.5 & 53.5 \\
PSCC-Net~\citep{psccnet}
& 7.8 & 5.5 & 14.2 & 58.8 & 42.4
& 11.3 & 9.0 & 14.0 & 64.4 & 53.1 \\
IML-ViT~\citep{imlvit}
& 15.8 & 11.3 & 27.9 & 50.7 & 43.7
& 6.0 & 4.7 & 1.9 & 76.0 & 50.1 \\
ObjectFormer~\citep{objectformer}
& 11.0 & 7.9 & 18.3 & 50.4 & 42.0
& 4.1 & 3.1 & 3.8 & 57.2 & 50.3 \\
TruFor~\citep{trufor}
& 16.5 & 11.9 & 18.1 & 69.5 & 43.0
& 23.8 & \underline{19.8} & 2.7 & \underline{79.5} & 50.5 \\
CAT-Net~\citep{catnet}
& \underline{33.1} & \underline{21.2} & \underline{45.3} & 51.7 & 47.7
& \underline{25.6} & 16.3 & N/A$^\dagger$ & 50.0 & 49.5 \\
DeCLIP~\citep{declip}
& 18.1 & 13.5 & 17.7 & 74.3 & 43.4
& 12.9 & 10.3 & 13.8 & 70.9 & 52.5 \\
MaskCLIP~\citep{opensdi}
& 8.0 & 5.9 & 36.3 & 73.1 & 51.1
& 10.7 & 9.0 & \underline{29.8} & 74.9 & \underline{58.1} \\
\midrule
\textbf{FUSED (Ours)}
& \textbf{51.4} & \textbf{38.9} & \textbf{86.2} & \textbf{89.9} & \textbf{83.1}
& \textbf{62.8} & \textbf{51.5} & \textbf{40.6} & \textbf{92.4} & \textbf{62.6} \\
\bottomrule
\end{tabular}
}
\caption{Zero-shot transfer of OpenSDID-trained models to the held-out AutoSplice and CocoGlide benchmarks. Dagger marks a degenerate predictor excluded from ranking. FUSED achieves the best result on every metric of both benchmarks.}
\label{tab:benchmarks_orig}
\end{table*}

\begin{table*}[!t]
\centering
{%
\small
\setlength{\tabcolsep}{4.0pt}
\begin{tabular}{@{}lcccccccccc@{}}
\toprule
\multirow{2}{*}{Model}
& \multicolumn{5}{c}{INP: Real vs.\ Inpaint}
& \multicolumn{5}{c}{INP-X: Real vs.\ Exchanged}
\\
\cmidrule(lr){2-6} \cmidrule(lr){7-11}
& Pix.\ F1 & IoU & F1 & AUC & Acc.
& Pix.\ F1 & IoU & F1 & AUC & Acc.
\\
\midrule
MVSS-Net~\citep{mvssnet}
& 10.6 & 7.5 & 61.0 & \underline{92.1} & 74.5
& 6.9 & 5.7 & 23.0 & 69.2 & 60.5 \\
PSCC-Net~\citep{psccnet}
& 15.3 & 10.4 & \underline{75.2} & 87.6 & \underline{79.3}
& 14.0 & 9.8 & 39.5 & 69.3 & 60.7 \\
IML-ViT~\citep{imlvit}
& 16.8 & 11.8 & 57.3 & 81.2 & 68.4
& 13.2 & 10.0 & 34.8 & 65.7 & 58.4 \\
ObjectFormer~\citep{objectformer}
& 6.2 & 4.4 & 17.3 & 63.3 & 58.4
& 5.7 & 4.2 & 12.8 & 60.0 & 57.1 \\
TruFor~\citep{trufor}
& 16.0 & 11.4 & 47.6 & 89.6 & 69.1
& 12.2 & 9.9 & 11.3 & 68.5 & 57.8 \\
CAT-Net~\citep{catnet}
& 13.2 & 8.1 & 39.2 & 43.2 & 45.4
& 13.0 & 8.0 & 39.7 & 44.4 & 45.7 \\
DeCLIP~\citep{declip}
& 12.9 & 9.6 & 18.7 & 68.4 & 59.3
& 14.2 & 11.2 & 13.9 & 65.6 & 58.0 \\
MaskCLIP~\citep{opensdi}
& \underline{17.1} & \textbf{13.1} & 61.3 & 80.7 & 70.5
& \underline{15.0} & \underline{12.5} & \underline{49.3} & \underline{74.0} & \underline{64.4} \\
\midrule
\textbf{FUSED (Ours)}
& \textbf{18.3} & \underline{12.5} & \textbf{84.4} & \textbf{93.8} & \textbf{80.8}
& \textbf{21.3} & \textbf{15.0} & \textbf{68.0} & \textbf{75.0} & \textbf{68.3} \\
\bottomrule
\end{tabular}
}
\caption{Zero-shot transfer of OpenSDID-trained models to the paired INP-X benchmark, with the global VAE artifact present (INP) and removed (INP-X). FUSED achieves the best detection accuracy and pixel F1 under both conditions.}
\label{tab:inpxtest}
\end{table*}

\subsection{Results}
\label{subsec:opensdi_results}

\paragraph{Cross-generator evaluation on OpenSDID.}
FUSED establishes a new state of the art on both tasks
(Tables~\ref{tab:opensdi_localization}
and~\ref{tab:opensdi_classification}). For localization, it achieves
the highest average IoU and pixel F1, $44.5$ and $55.3$, compared with
$42.7$ and $49.4$ for MaskCLIP~\citep{opensdi}.

The advantage grows with the generator gap. On unseen SDXL, SD3, and
Flux.1, FUSED improves pixel F1 by $11.3$, $10.9$, and $11.5$ points.
On in-domain SD1.5, MaskCLIP and CAT-Net remain stronger, while
MaskCLIP obtains the best SD2.1 IoU, $55.5$ versus $52.4$.

For detection, FUSED achieves an average F1 of $88.0$ and accuracy of
$89.5$, compared with $77.8$ and $82.0$ for MaskCLIP. The F1 margins
reach $16.1$ points on SD3 and $14.7$ on Flux.1. Section~D of the
supplementary material shows that FUSED leads MaskCLIP in every edit-size
bin, while the failures in Section~F occur mostly in the smallest bins.
Overall, FUSED improves cross-generator transfer by adaptively balancing
forensic and semantic signals.

\paragraph{Transfer to held-out benchmarks.}
\label{subsec:transfer_results}
FUSED reaches $51.4$ pixel F1 on AutoSplice and $62.8$ on CocoGlide
(Table~\ref{tab:benchmarks_orig}), improving over CAT-Net, the best
previous method, by $18.3$ and $37.2$ points, while several baselines
drop to single-digit scores.
Detection under this shift suffers from a threshold mismatch: the
score distributions move away from the fixed threshold of $0.5$. The
baselines still tend to rank real below fake, with ROC-AUC up to
$79.5$, yet their AutoSplice accuracy falls to near or below chance
($42.0$--$53.7$).
FUSED obtains the highest ROC-AUC on both benchmarks ($89.9$ and
$92.4$) and remains usable at the fixed threshold on AutoSplice
($83.1$ accuracy). On CocoGlide the mismatch also affects FUSED: its
accuracy of $62.6$ is the highest among all methods but below
its ROC-AUC. CAT-Net's nominally best CocoGlide detection F1 comes
from a degenerate predictor with chance-level ROC-AUC and is excluded from ranking.

\paragraph{Artifact reliance on the INP-X pair.}
The paired conditions in Table~\ref{tab:inpxtest} directly test
whether decisions rest on the manipulated region or on the global
artifact. FUSED obtains the highest accuracy and pixel F1 under both
conditions. Removing the artifact costs it $12.5$ accuracy points
($80.8$ to $68.3$), compared with $18.6$ for PSCC-Net, the second-most
accurate method under INP. MaskCLIP loses only $6.1$ points but never
exceeds $70.5$. Every method above chance is more accurate with the artifact present, so none of the evaluated methods decides fully within the manipulated region. However, FUSED
remains the strongest under both conditions.

\begin{table*}[!t]
\centering
{%
\small
\setlength{\tabcolsep}{4.0pt}
\begin{tabular}{@{}l*{6}{r}@{\hspace{1.2em}}*{6}{r}@{}}
\toprule
\multirow{2}{*}{Model}
& \multicolumn{6}{c}{Localization: pixel F1}
& \multicolumn{6}{c}{Detection: F1}
\\
\cmidrule(lr){2-7} \cmidrule(lr){8-13}
& SD1.5 & SD2.1 & SDXL & SD3 & Flux.1 & Avg.
& SD1.5 & SD2.1 & SDXL & SD3 & Flux.1 & Avg.
\\
\midrule
\textbf{FUSED (Ours)}
& \textbf{70.5} & \textbf{63.9} & \textbf{48.3} & \textbf{62.1} & \textbf{31.8} & \textbf{55.3}
& 93.9 & \textbf{95.4} & \textbf{90.2} & \textbf{89.2} & \underline{71.2} & \textbf{88.0} \\
\midrule
w/o MoE (dense FFN)
& \underline{70.2} & \underline{63.8} & \underline{47.2} & 60.6 & 29.0 & 54.2
& 94.7 & 91.0 & 81.5 & 80.8 & 62.4 & 82.1 \\
w/o seg-bridge
& 57.9 & 44.6 & 27.0 & 35.0 & 16.8 & 36.2
& \underline{96.0} & 91.8 & 84.9 & 86.2 & 70.2 & 85.8 \\
w/o ConvNeXt
& 29.3 & 19.9 & 16.0 & 17.7 & 12.8 & 19.1
& 86.7 & 83.1 & 66.3 & 76.3 & 70.5 & 76.6 \\
w/o SparseViT
& 61.6 & 60.5 & 39.2 & 58.1 & 28.1 & 49.5
& \textbf{96.2} & 92.6 & 83.1 & 83.0 & 63.4 & 83.7 \\
\midrule
ConvNeXt-Large
& 68.0 & 62.0 & 42.7 & 56.3 & 26.0 & 51.0
& 93.7 & 92.2 & 82.8 & 83.6 & 64.5 & 83.4 \\
ConvNeXt-Base
& 65.4 & 54.6 & 37.4 & 48.0 & 24.4 & 46.0
& 91.3 & 87.8 & 71.1 & 74.9 & 57.6 & 76.5 \\
\midrule
MoE top-$k{=}1$
& 67.0 & 61.9 & 44.0 & 58.7 & 26.8 & 51.7
& 94.2 & \underline{94.1} & \underline{89.5} & \underline{89.1} & \textbf{71.8} & \underline{87.8} \\
MoE top-$k{=}3$
& 70.0 & \underline{63.8} & 47.0 & \underline{61.3} & \underline{29.8} & \underline{54.4}
& 95.8 & 92.4 & 86.4 & 86.6 & 67.7 & 85.8 \\
\bottomrule
\end{tabular}
}
\caption{Ablation of FUSED on the OpenSDID test generators (pixel F1
and detection F1). The blocks remove one component at a time, vary the
frozen semantic backbone, and vary the routing sparsity. Every variant
is retrained on the SD1.5 split with the same recipe, and the FUSED
row (ConvNeXt-XXL, top-$k{=}2$) is the shared reference. Bold marks
the best and the underlined are second-best value in each column across
all rows. FUSED achieves the best average on both tasks.}
\label{tab:abl_all}
\end{table*}

\begin{table}[H]
\centering
{%
\small
\setlength{\tabcolsep}{4.0pt}
\begin{tabular}{@{}lccccc@{}}
\toprule
Model & Pix.\ F1 & IoU & F1 & AUC & Acc. \\
\midrule
MaskCLIP
& 28.3 & 20.5 & 87.7 & 89.3 & 82.4 \\
\textbf{FUSED (Ours)}
& \textbf{42.8} & \textbf{33.6} & \textbf{89.2} & \textbf{89.7} & \textbf{83.9} \\
\bottomrule
\end{tabular}
}
\caption{Transfer to So-Fake-OOD after training on the multi-generator
So-Fake-Set. FUSED improves on every metric, with the largest gains in
localization.}
\label{tab:sofake}
\end{table}

\paragraph{Large-scale OOD transfer on So-Fake.}
Table~\ref{tab:sofake} reports transfer to So-Fake-OOD after training
on So-Fake-Set. FUSED improves on every metric. The image-level
detection gains are modest, within $1.5$ points, while the localization
gains are substantially larger: $+14.5$ points in pixel F1 and $+13.1$
points in IoU. FUSED therefore remains ahead on both tasks when trained on a heterogeneous multi-generator source and evaluated on held-out contemporary commercial generators. Section~G of
the supplementary material provides qualitative examples.

\paragraph{Qualitative Analysis.}
\label{subsec:qualitative}Figure~\ref{fig:qualitative_opensdi} compares predicted masks across
four unseen OpenSDID generators. Under generator shifts, forensic baselines fail first: MVSS-Net, TruFor, CAT-Net, and ObjectFormer return empty or near-empty masks, whereas PSCC-Net and IML-ViT produce fragmented predictions. The CLIP-based MaskCLIP and DeCLIP keep
producing coherent masks, but these are coarse and often extend
beyond the edit. FUSED remains closest to the ground truth,
recovering the overall shape without collapse. Sections~A--C of the supplementary material extend this comparison to further OpenSDID,
fully synthetic, and small-edit examples.

\subsection{Ablation Studies}
\label{subsec:ablation}


\begin{figure}[!tb]
    \centering
    \includegraphics[width=\linewidth]{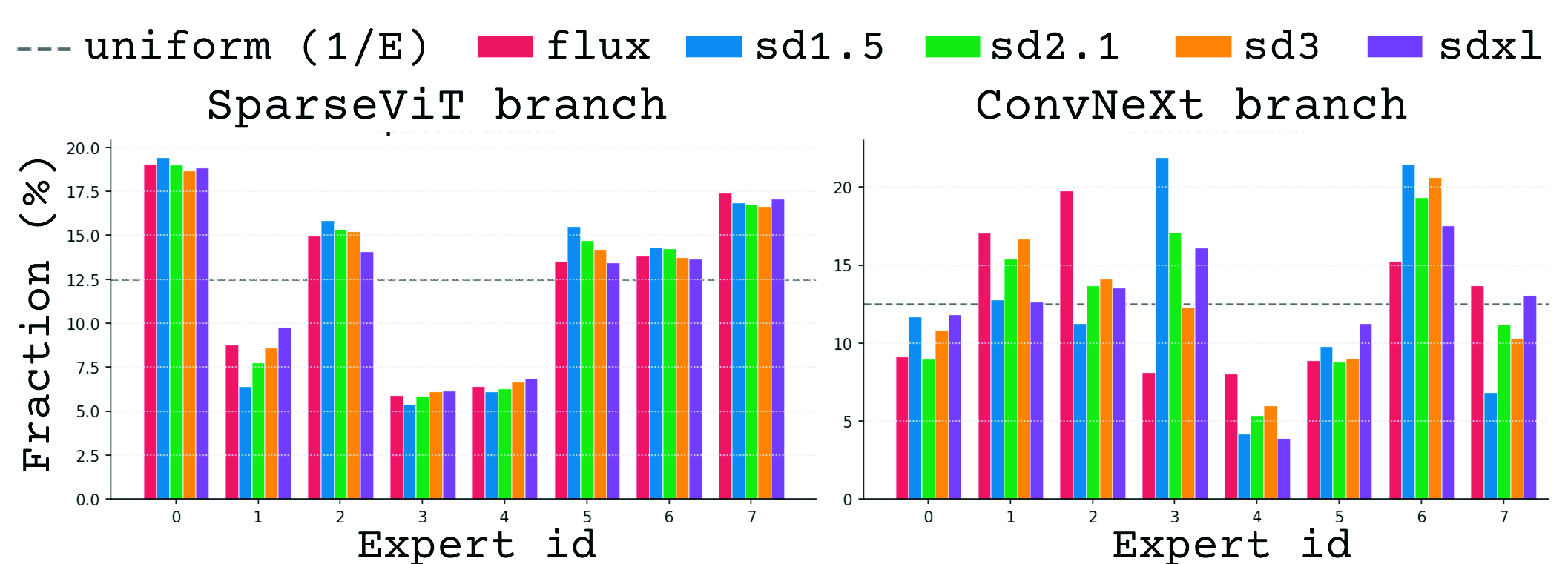}
    \caption{Per-token utilization of the MoE experts
    ($E{=}8$, top-$k{=}2$) for the SparseViT token group on the left
    and the ConvNeXt token group on the right. Each bar series
    corresponds to one generator, and the dashed line denotes uniform
    utilization, $1/E$. All experts remain active, while utilization
    differs between branches and varies across generators.}
    \label{fig:routing}
\end{figure}

\paragraph{Component ablation.}
We remove each of the four components of FUSED individually, retrain
each resulting variant on the SD1.5 split using the same training
recipe, and evaluate all variants on every generator
(Table~\ref{tab:abl_all}). Removing the frozen CLIP-ConvNeXt
branch causes the largest decrease on both tasks, reducing the average
localization pixel F1 to $19.1$ and detection F1 to $76.6$. This
confirms that the semantic branch provides an important foundation for
both tasks.

Removing the segmentation bridge reduces average localization pixel F1
from $55.3$ to $36.2$, while affecting detection less severely.
Removing the forensic SparseViT branch reduces localization on every
generator, including the in-domain SD1.5 split, where pixel F1 decreases
from $70.5$ to $61.6$, and lowers the average from $55.3$ to $49.5$.

The MoE fusion has a more transfer-specific effect. To isolate conditional routing from the capacity it activates, we replace the
$E{=}8$, top-$k{=}2$ mixture with a single SwiGLU FFN whose hidden width matches
the two experts active per token, leaving the active parameter count unchanged. This changes localization only slightly, from $55.3$ to $54.2$, and leaves in-domain detection almost unchanged. However, it reduces detection F1 by $4.4$ to $8.8$ points on each unseen
generator, a consistent loss across all four transfer settings. The gain, therefore, comes from routing rather than from the capacity it activates. Qualitative masks for each ablated variant are shown in Section~E of the
supplementary material.

\paragraph{Routing behavior.}
Figure~\ref{fig:routing} presents the per-token expert utilization of
both branches across all five generators. The forensic and semantic
token groups exhibit distinct utilization profiles, which vary
across generators, and the
load-balancing loss keeps all eight experts active in both branches.
The router thus treats the forensic and semantic streams differently
and adapts the expert mixture to each input. Varying the routing
sparsity (Table~\ref{tab:abl_all}) shows that the number of active
experts matters: top-$k{=}1$ nearly matches FUSED on
detection, with marginally higher Flux.1 F1, but loses $3.6$ average
localization points, while top-$k{=}3$ recovers localization yet
drops detection F1 on every unseen generator. Top-$k{=}2$ therefore
gives the best trade-off between the two tasks.

\paragraph{Backbone capacity.}
Increasing the capacity of the frozen semantic backbone improves both
tasks monotonically (Table~\ref{tab:abl_all}). As the backbone is
scaled from ConvNeXt-Base to ConvNeXt-Large and then to
ConvNeXt-XXL, average pixel F1 increases from $46.0$ to $51.0$ and
$55.3$, respectively. Detection F1 similarly increases from $76.5$ to
$83.4$ and $88.0$. Semantic-backbone capacity therefore has a direct
effect on cross-generator transfer performance.

\begin{figure}[t]
    \centering
    \includegraphics[width=1.0\linewidth]{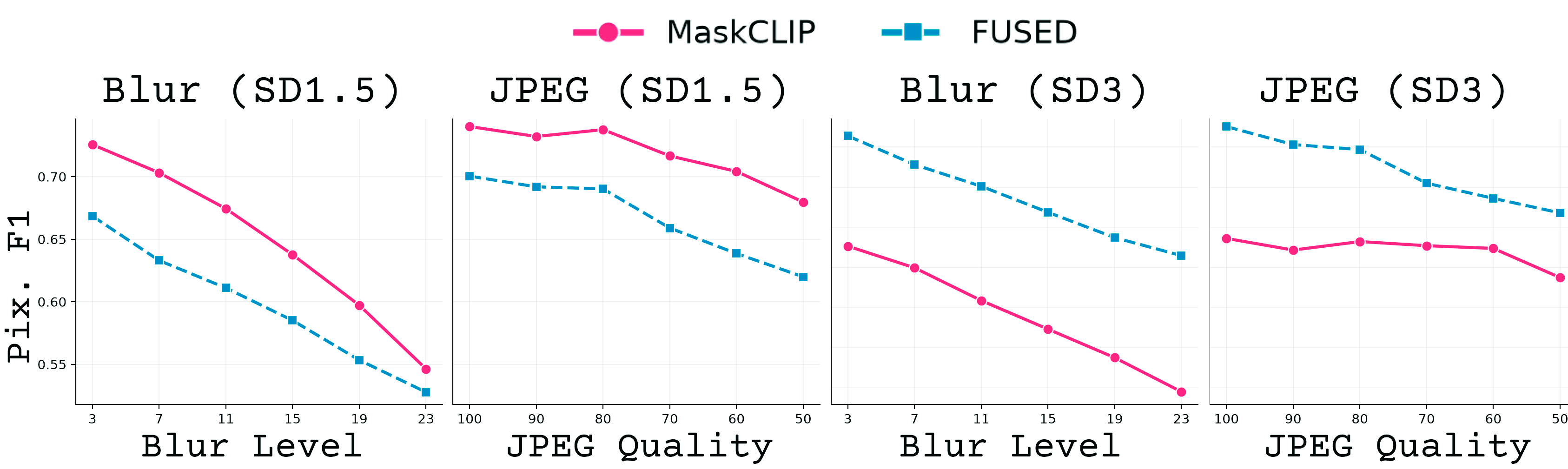}
    \caption{Pixel-F1 robustness to Gaussian blur and JPEG compression on in-domain SD1.5 and unseen SD3. FUSED maintains its SD3 advantage across all severity levels, showing that its unseen-generator transfer remains robust even when the input is degraded.}
    \label{fig:robustness}
\end{figure}

\paragraph{Robustness to post-processing.}
We evaluate localization robustness under Gaussian blur and JPEG
compression, two common post-processing operations that may suppress
the low-level traces used by forensic detectors
(Figure~\ref{fig:robustness}). Neither perturbation changes the overall
ranking. MaskCLIP retains its advantage on the in-domain SD1.5 split,
whereas FUSED remains stronger on the unseen SD3 split across all
evaluated severity levels. The cross-generator advantage of FUSED, therefore, does not depend on pristine inputs.

\section{Conclusion}
We introduce FUSED, a unified model for AI inpainting detection and
localization that integrates forensic and semantic representations through
a token-wise sparse Mixture-of-Experts followed by cross-stream attention.
A segmentation bridge propagates localization supervision into the fused
representation. Experiments on OpenSDID, AutoSplice, CocoGlide, and INP-X
demonstrate state-of-the-art transfer from a single training source, with the
largest gains on unseen generators and the strongest overall performance
both with and without the global VAE artifact. When retrained on the
multi-generator So-Fake-Set, FUSED remains ahead of the strongest
baseline on So-Fake-OOD in both detection and localization. This shows that its advantage is not tied to single-generator training and extends to a contemporary benchmark containing held-out commercial generators.

\textbf{Limitations.} FUSED remains the strongest after removing the
global VAE artifact, yet its accuracy drop on INP-X suggests residual
sensitivity to this cue. Localization is harder for the smallest edits,
although FUSED retains its advantage across all edit-size bins, as shown in
the supplementary material. Finally, our experiments focus on
generation-based inpainting, leaving other manipulation types for future
work.

\bibliography{aaai2027}
\clearpage
\appendix
%

\frenchspacing  
\pdfinfo{
/TemplateVersion (2027.1)
}
\setcounter{secnumdepth}{2}

\twocolumn[
\begin{center}
{\LARGE\bfseries
FUSED: Forensic--Semantic Mixture-of-Experts for AI Inpainting\\
Detection and Localization \\
(Supplementary Material)
\par}
\vspace{1.5em}
\end{center}
]
\appendix
The supplementary material extends the evaluation in the main paper with
additional qualitative and stratified analyses. All results use the same
checkpoints and baselines evaluated in Section~4 of the main paper.
Sections~\ref{sec:supp_opensdid}--\ref{sec:supp_small} examine localization
behavior on representative OpenSDID examples, fully synthetic images, and
images with small manipulated regions. Section~\ref{sec:supp_area}
stratifies localization and detection by manipulated-area fraction,
Section~\ref{sec:supp_ablation} visualizes the component ablations,
Section~\ref{sec:supp_failure} examines errors from both output heads, and
Section~\ref{sec:supp_sofake} provides qualitative results under the
So-Fake-Set to So-Fake-OOD distribution shift. Together, these analyses
clarify where the gains in the main paper arise and which cases remain
challenging.

\section{Additional Localization Examples}
\label{sec:supp_opensdid}
Figure~\ref{fig:supp_opensdid} shows ten randomly selected OpenSDID test
examples. MVSS-Net produces spatially diffuse predictions in most rows,
while several baselines frequently return empty or nearly
empty masks. MaskCLIP, the strongest localization baseline, recovers the coarse
manipulated region in several examples but remains inconsistent and produces
empty predictions in others. FUSED responds across all examples and usually
preserves the coarse extent of the manipulated region, although its masks
sometimes include neighboring objects or background areas. Overall, FUSED's
main qualitative advantage is more consistent localization.

\section{Fully Synthetic Images}
\label{sec:supp_fake}
\begin{figure*}[t]
    \centering
    \includegraphics[width=\textwidth,keepaspectratio]{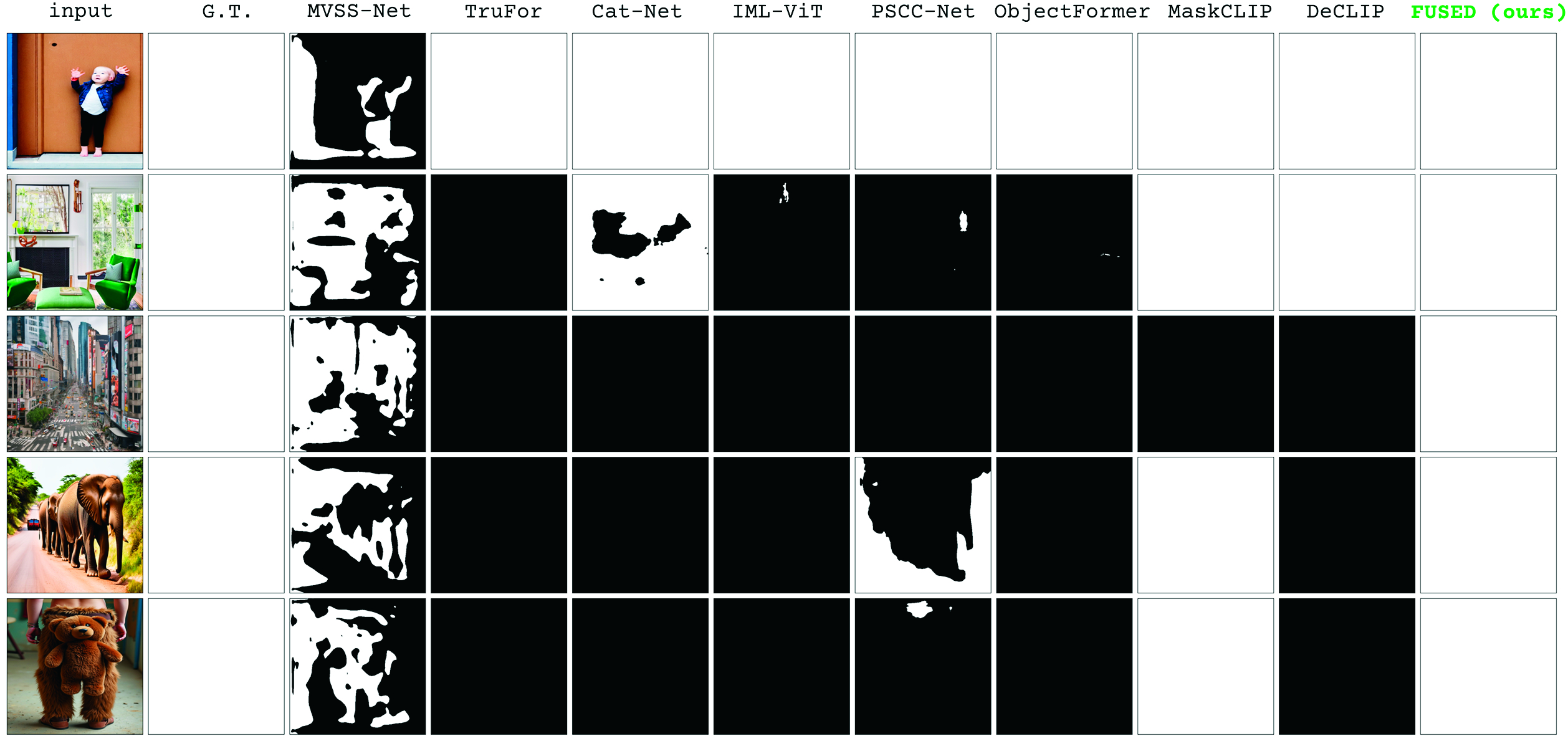}
    \caption{Qualitative results on fully synthetic images.
    The ground-truth mask covers the entire image because every pixel is
    generated. FUSED predicts a full-frame mask in all five examples,
    whereas the compared baselines are less consistent.}
    \label{fig:supp_fake}
\end{figure*}

Figure~\ref{fig:supp_fake} presents five fully synthetic images. Because
every pixel is generated, the correct mask covers the entire frame. FUSED
produces a full-frame prediction in all five examples. MaskCLIP and DeCLIP
also produce full-frame masks in several cases, whereas the forensic
baselines frequently return empty or incomplete predictions.

These examples show that FUSED can extend its response across a fully
generated image even when there is no localized authentic--generated
boundary.

\section{Small-Manipulation Examples}
\label{sec:supp_small}
\begin{figure*}[t]
    \centering
    \includegraphics[width=\textwidth,keepaspectratio]{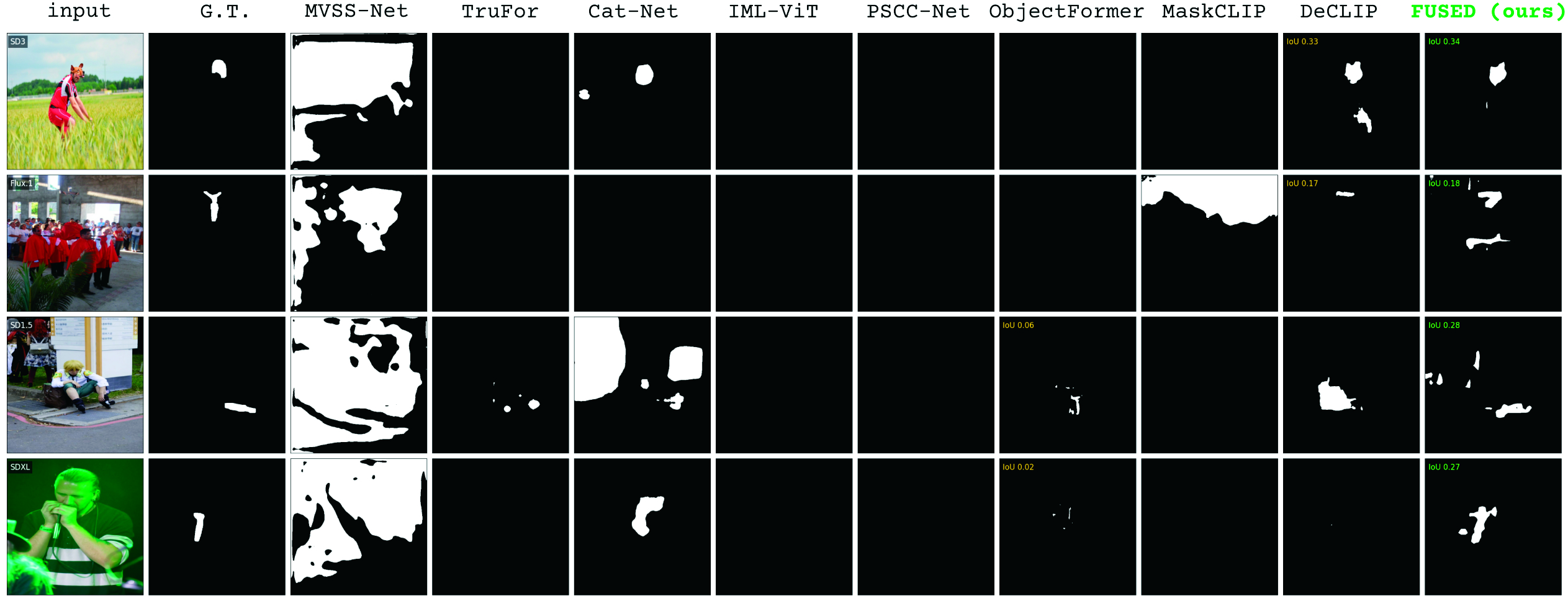}
    \caption{Qualitative results for small manipulated regions.
The edited region occupies only a small fraction of each image.
FUSED generally localizes the target region, although its predictions may
include additional disconnected areas.}
    \label{fig:supp_small}
\end{figure*}

Figure~\ref{fig:supp_small} shows examples in which the edited region
occupies only a small fraction of the image. MVSS-Net substantially
over-segments every example, while many other baselines return empty masks.
Overall, FUSED produces non-empty, localized predictions in all four cases,
although some masks contain spurious or fragmented components.

\begin{figure}[!tb]
    \centering
    \includegraphics[width=\linewidth,keepaspectratio]{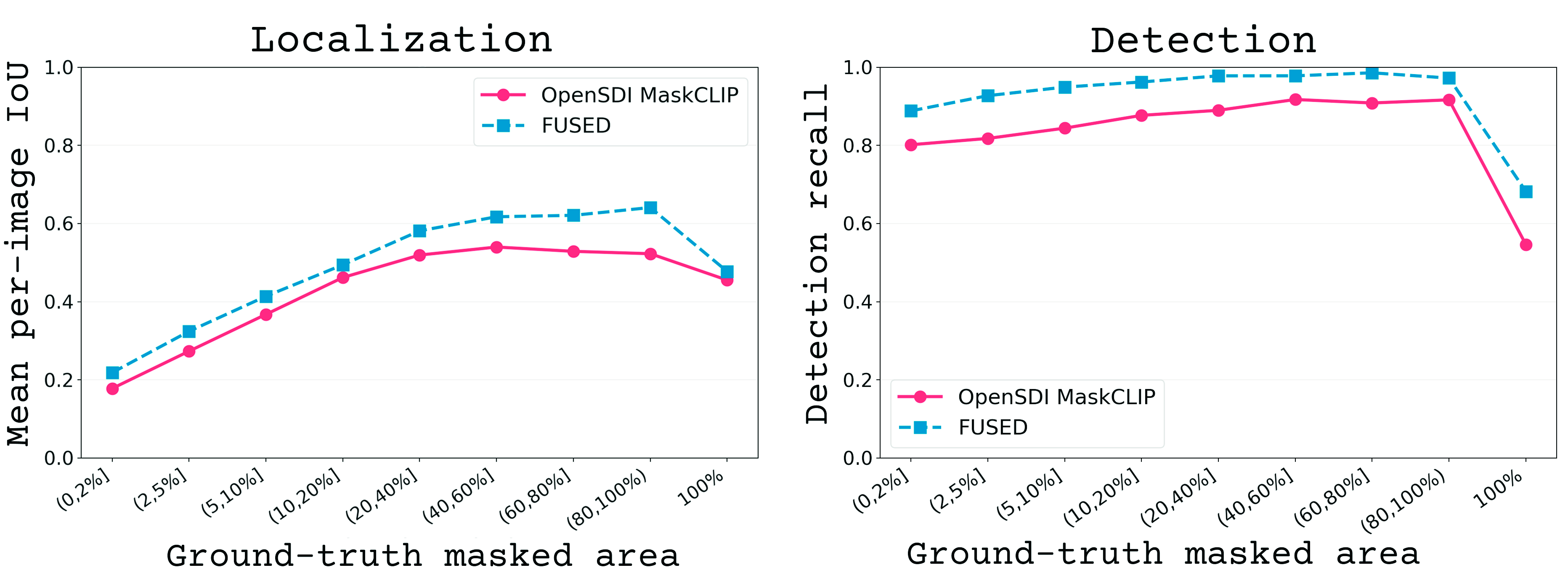}
    \caption{Performance by manipulated-area fraction on the
    OpenSDID test set. Mean per-image localization IoU (left) and
    detection recall on manipulated images (right), binned by the
    ground-truth manipulated-area fraction. FUSED exceeds MaskCLIP in
    every bin on both tasks. Localization improves from the smallest to
    the medium and large partial-edit bins, while recall remains high
    across partial edits.}
    \label{fig:supp_area}
\end{figure}

\section{Performance by Manipulated-Area Fraction}
\label{sec:supp_area}
Figure~\ref{fig:supp_area} reports mean per-image localization IoU and
detection recall across bins of ground-truth manipulated-area fraction.
We compare FUSED with MaskCLIP, the strongest localization baseline in
Table~1 of the main paper. FUSED outperforms MaskCLIP in every bin on both tasks, showing that its
aggregate advantage is not driven by one manipulation scale. Localization
IoU increases from below $0.25$ for edits covering less than $2\%$ of the
image to above $0.60$ in the medium and large partial-edit bins. Detection
recall is considerably less sensitive to edit size and remains above
approximately $0.90$.

Both methods decline on fully generated images, which
form a distinct setting without an authentic--generated boundary and are
shown qualitatively in Section~\ref{sec:supp_fake}. Overall, edit size affects localization more than detection: small edits may be detected but localized imprecisely.

\begin{figure}[!tb]
    \centering
    \includegraphics[width=\linewidth,keepaspectratio]{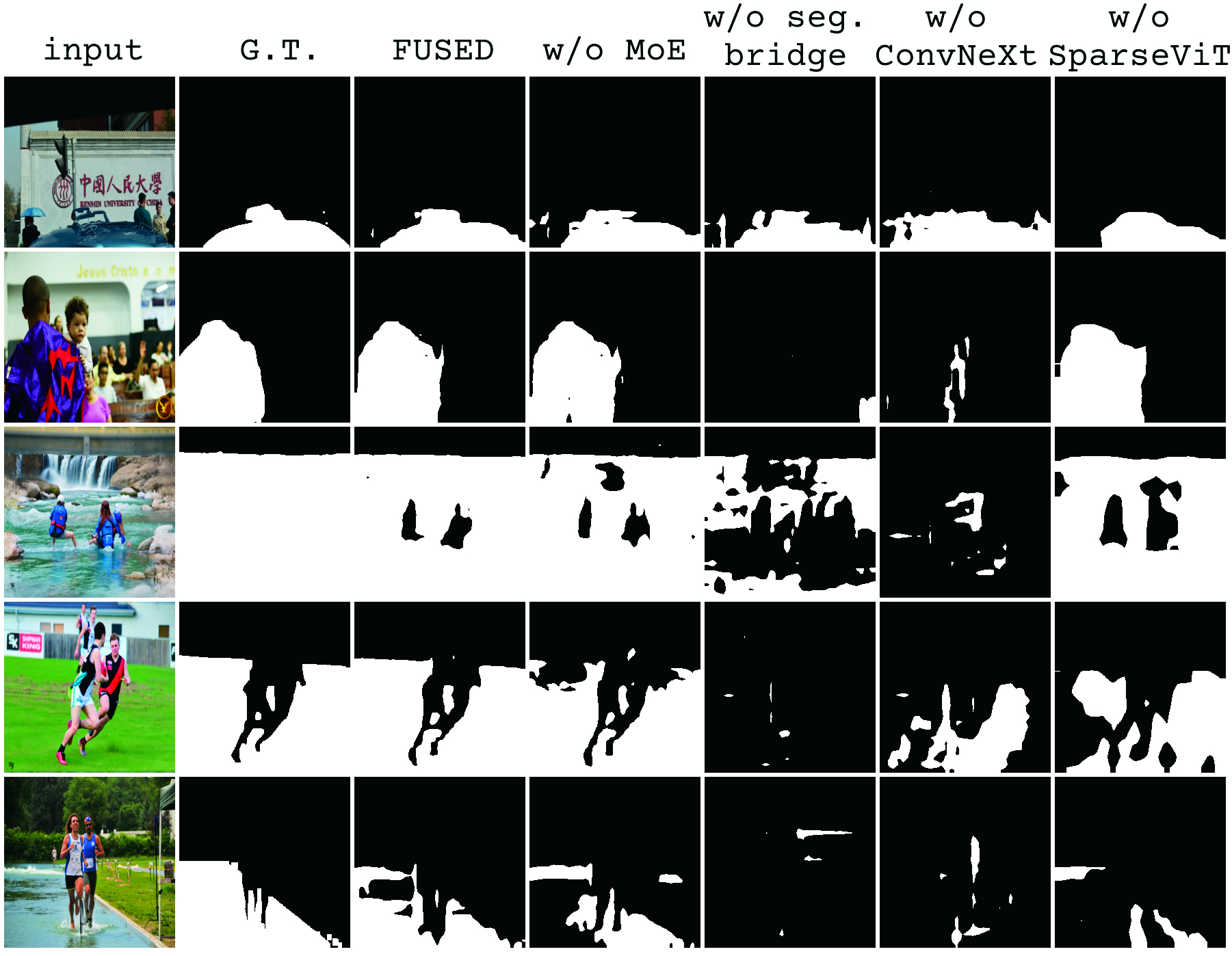}
    \caption{Qualitative ablation results. Predictions of the
    full model and the four component-ablation variants from Table~5 of
    the main paper on OpenSDID test examples. Removing the segmentation
    bridge, SparseViT or ConvNeXt branch produces the largest qualitative changes.}
    \label{fig:quali_ablation}
\end{figure}

\section{Qualitative Ablation Results}
\label{sec:supp_ablation}
Figure~\ref{fig:quali_ablation} complements the component ablation shown
in Table~5 of the main paper with the predicted masks themselves, comparing the
full model against the four ablated variants on OpenSDID test samples. The
degradation visible in the masks tracks the quantitative drops. Removing the
segmentation bridge, SparseViT or the ConvNeXt branch damages the predictions most,
yielding fragmented or displaced regions where the full model recovers a
coherent shape.

\begin{figure}[!tb]
    \centering
    \includegraphics[width=\linewidth,keepaspectratio]{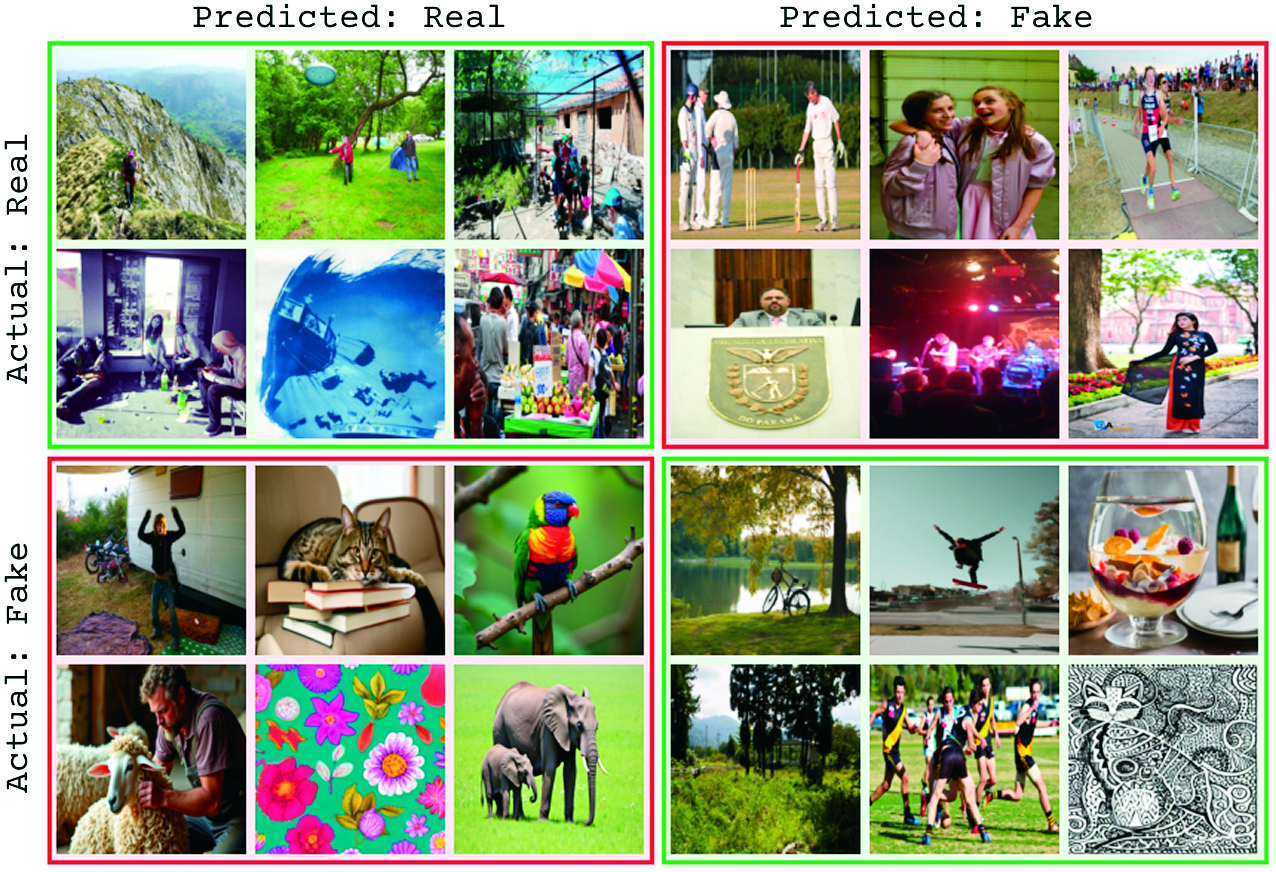}
    \caption{Detection outcomes on the OpenSDID test set.
    Images are grouped by their actual and predicted real/fake labels. The off-diagonal cells show real
    images classified as fake and fake images classified as real.}
    \label{fig:supp_fail_det}
\end{figure}

\section{Failure Cases}
\label{sec:supp_failure}
The preceding sections show representative and stratified behavior. Here, we
examine cases in which FUSED fails, considering the detection and localization
heads separately.

\subsection{Detection Errors}
\label{subsec:supp_fail_det}
Figure~\ref{fig:supp_fail_det} groups test images by their actual and
predicted image-level labels. The false-positive and false-negative examples
cover varied content, including people, sporting events, animals, isolated
objects, natural scenes, and graphic patterns. No single content category or
visual characteristic explains all errors in either direction. 

The examples, therefore, suggest that the remaining detection errors are
heterogeneous rather than dominated by one obvious semantic shortcut.

\begin{figure}[!tb]
    \centering
    \includegraphics[width=\linewidth,keepaspectratio]{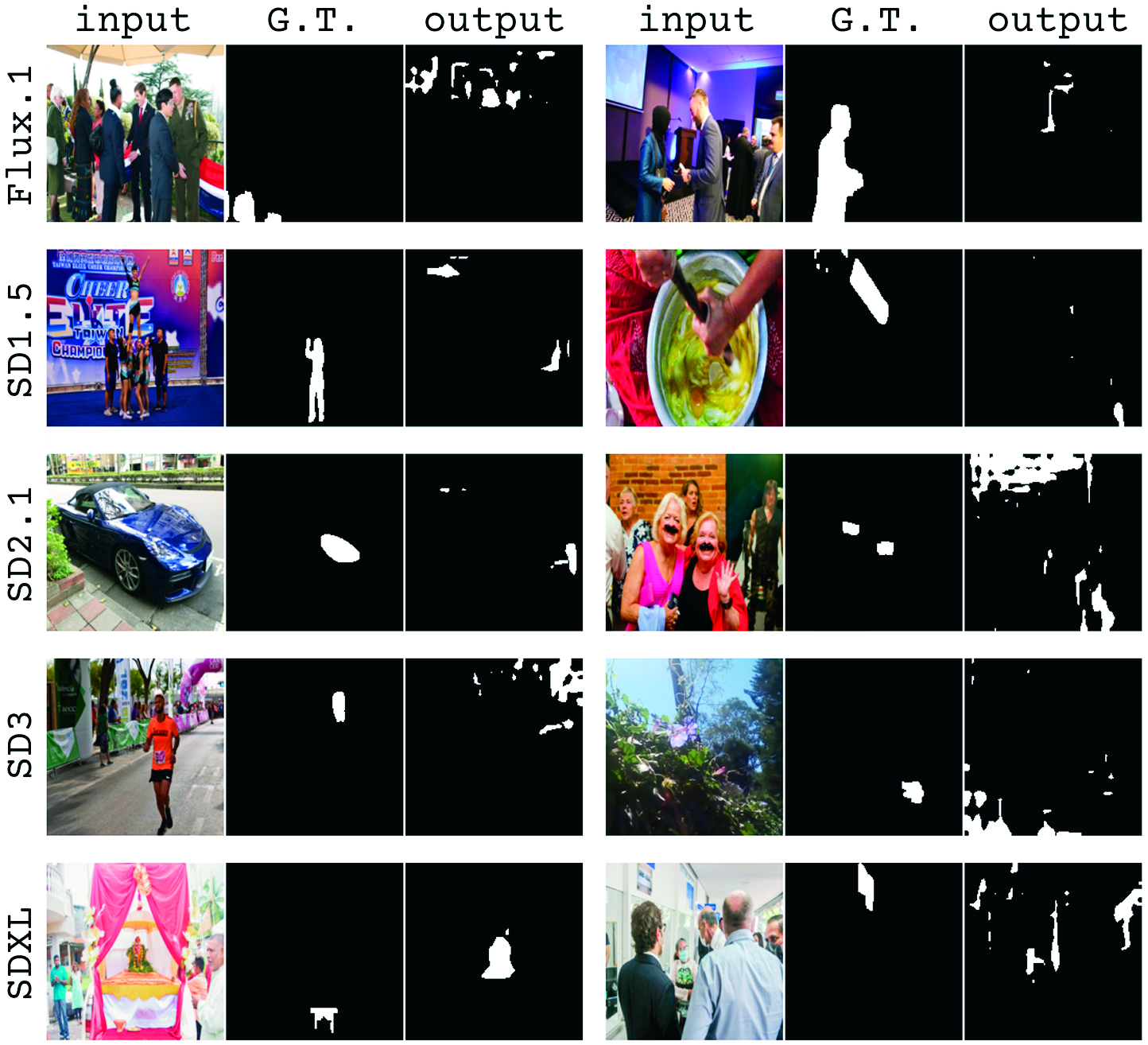}
    \caption{Ten OpenSDID examples,
    two per generator, for which FUSED obtains an IoU of $0.00$.
    Each triplet contains the input image, ground-truth mask, and
    predicted mask.}
    \label{fig:supp_fail_loc}
\end{figure}

\subsection{Localization Failures}
\label{subsec:supp_fail_loc}
Figure~\ref{fig:supp_fail_loc} shows ten examples, two per generator, for
which FUSED obtains an IoU of $0.00$. The ground-truth regions are generally
small relative to the image, placing these examples in the lowest-area regime
of Figure~\ref{fig:supp_area}.

\begin{figure*}[t]
    \centering
    \includegraphics[width=\textwidth,keepaspectratio]{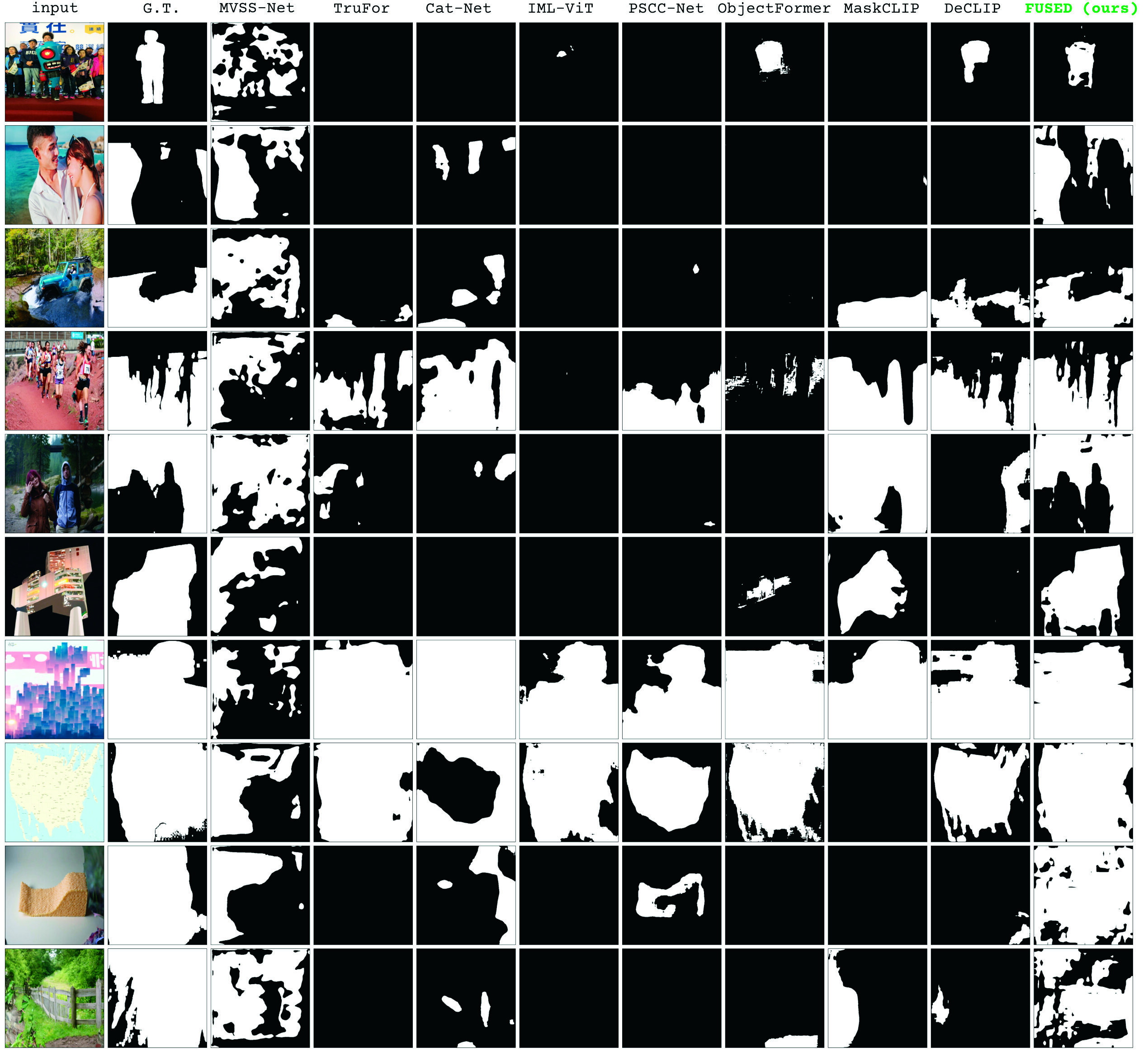}
    \caption{Additional qualitative results on OpenSDID.
    Ten randomly selected test examples. Each row contains the input image,
    ground-truth mask, and predictions from all compared methods. FUSED
    responds across all examples and generally recovers the coarse
    manipulated region, although some predictions remain over-segmented.}
    \label{fig:supp_opensdid}
\end{figure*}

Most predicted masks are not empty. Instead, FUSED responds to regions
elsewhere in the image, sometimes with an area comparable to or larger than
the target. The zero IoU, therefore, reflects incorrect spatial
attribution rather than mask collapse. This differs from the
frequent empty-mask behavior of several baselines in
Figure~\ref{fig:supp_opensdid} and indicates that improving these cases
requires more precise spatial grounding rather than simply stronger mask
activation.

\section{So-Fake-OOD Qualitative Results}
\label{sec:supp_sofake}

\begin{figure*}[t]
    \centering
    \includegraphics[width=\textwidth,keepaspectratio]{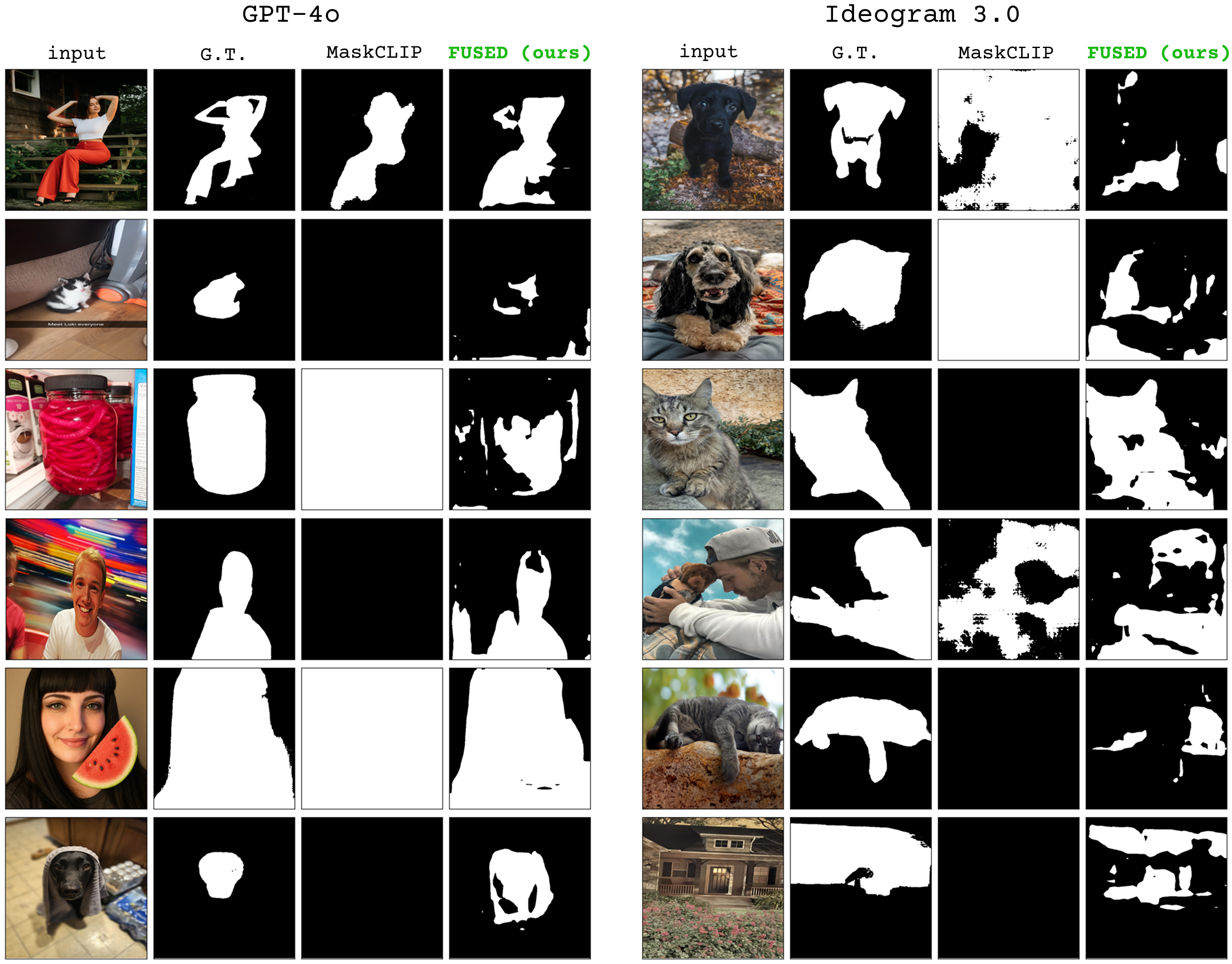}
    \caption{Qualitative localization results on So-Fake-OOD for GPT-4o
    (left) and Ideogram~3.0 (right). Each group shows the input image,
    ground-truth mask, MaskCLIP prediction, and FUSED prediction. MaskCLIP
    frequently produces empty or near-full masks, whereas FUSED more often
    localizes the manipulated region, although some predictions remain
    fragmented or over-segmented.}
    \label{fig:supp_sofake}
\end{figure*}

Figure~\ref{fig:supp_sofake} complements the quantitative So-Fake-OOD
results in the main paper with examples from GPT-4o and Ideogram~3.0.
MaskCLIP frequently produces empty or near-full predictions, whereas FUSED
more often produces localized responses that overlap the manipulated region.
The task nevertheless remains challenging: several FUSED predictions contain
fragmented components or extend beyond the ground-truth region. Overall, the
examples are consistent with the quantitative results, where the largest
advantage of FUSED over MaskCLIP is in localization rather than image-level
detection.


\end{document}